\documentclass[runningheads]{llncs}
\usepackage{url}
\usepackage{cite}
\usepackage{amsmath}
\usepackage{hyperref}
\usepackage{graphicx}
\usepackage{subcaption}
\usepackage{multirow}
\usepackage{booktabs}
\usepackage{xcolor}
\usepackage{wrapfig}

\let\svthefootnote\thefootnote
\newcommand\freefootnote[1]{%
  \let\thefootnote\relax%
  \footnotetext{#1}%
  \let\thefootnote\svthefootnote%
}

\newcommand{\DOI}[1]{\textbf{DOI}: #1}

\begin{document}

\title{Sidewalk Measurements from Satellite Images: Preliminary Findings}
\titlerunning{SDSS 2021}

\author{Maryam Hosseini\inst{1,4,*} \and Iago B. Araujo \inst{2} \\ \and Hamed~Yazdanpanah\inst{2} \and Eric K. Tokuda\inst{2} \and\\Fabio Miranda\inst{3} \and Claudio T. Silva\inst{1} \and Roberto~M. Cesar~Jr.\inst{2}}
\institute{New York University (NYU), New York, NY, United States\\\and University of São Paulo (USP), São Paulo, SP, Brazil\\\and University of Illinois at Chicago (UIC), Chicago, IL, United States\\\and Rutgers University, Newark, NJ, United States}

\authorrunning{M. Hosseini et al.}
\maketitle              % typeset the header of the contribution

\begin{abstract}
Large-scale analysis of pedestrian infrastructures, particularly sidewalks, is critical to human-centric urban planning and design.
Benefiting from the rich data set of planimetric features and high-resolution orthoimages provided through the New York City Open Data portal, we train a computer vision model to detect sidewalks, roads, and buildings from remote-sensing imagery and achieve 83\% mIoU over held-out test set. We apply shape analysis techniques to study different attributes of the extracted sidewalks. More specifically, we do a tile-wise analysis of the width, angle, and curvature of sidewalks, which aside from their general impacts on walkability and accessibility of urban areas, are known to have significant roles in the mobility of wheelchair users. The preliminary results are promising, glimpsing the potential of the proposed approach to be adopted in different cities, enabling researchers and practitioners to have a more vivid picture of the pedestrian realm. 

\keywords{sidewalks \and semantic segmentation \and urban analytics \and accessibility.}
\end{abstract}
\DOI{\url{https://doi.org/10.25436/E2QG6F}}
% \vspace{-0.75cm}
\section{Introduction} \label{sec:intro}
\vspace{-0.3cm}

\freefootnote{* Corresponding author: \email{maryam.hosseini@nyu.edu}}

Pedestrian infrastructure has a significant impact on the everyday life of people, specifically those with special needs, for whom such infrastructures are the primary means of accessing public spaces~\cite{saha2019project}.
The presence, condition, width, and shape of sidewalks are shown to impact pedestrians' safety and accessibility~\cite{osama2017evaluating}. The curvature is one of the critical factors in ensuring safe navigation for wheelchair users~\cite{beale2006mapping}. 
\begin{figure}[t!]
\centering
	\includegraphics[width=.8\linewidth]{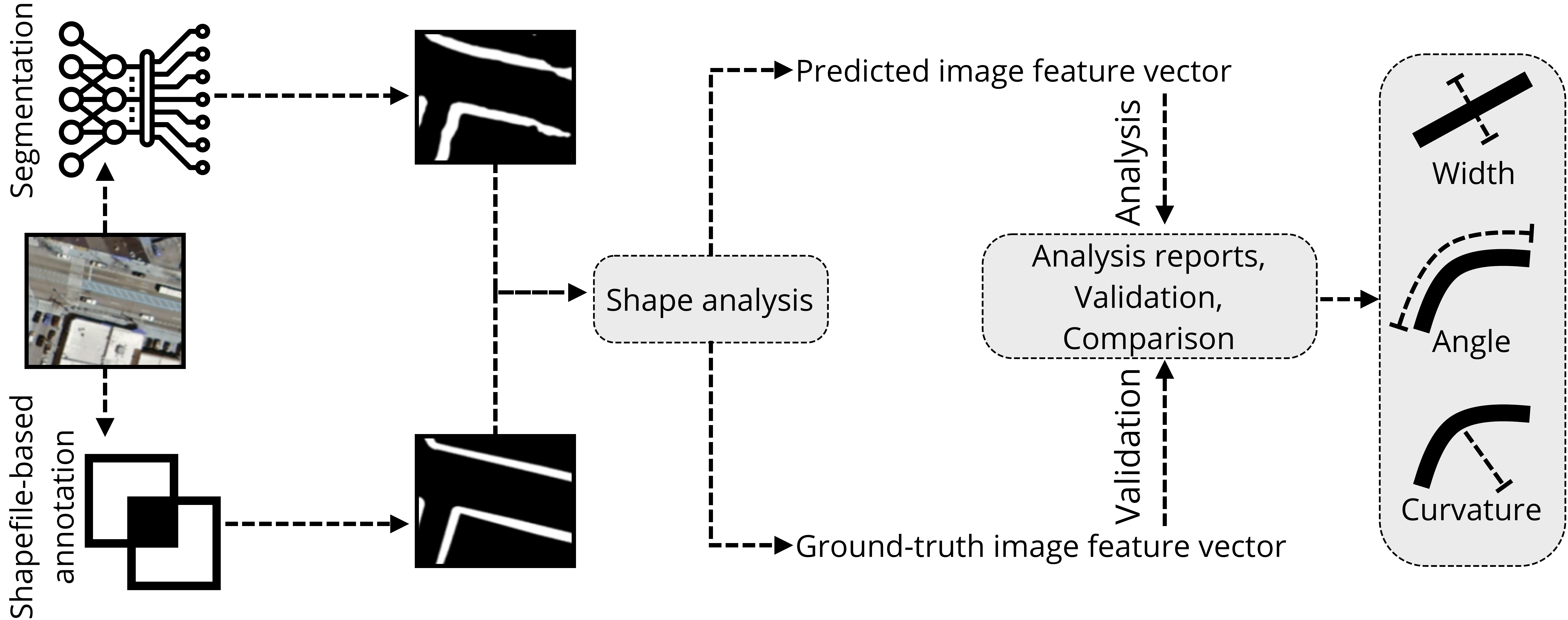}
	\caption{Proposed method with both analysis and validation components. \label{fig:pipeline}}
\vspace{-0.5cm}
\end{figure}
Despite their importance, there is a significant lack of city-wide fine-level sidewalk data, which poses challenges to the assessment and planning of pedestrian infrastructures~\cite{deitz2021squeaky}. The time and cost-intensive nature of in-field data collection have for long been a limiting factor in expanding the research on the built environment~\cite{rundle2011using}. The availability of new sources of data such as aerial and street-level images, together with the advent of new computer vision techniques, opened new frontiers in measuring the physical form of the cities.~\cite{saha2019project,mosaic}. 
Sidewalks, however, have been noticeably overlooked. The majority of the sidewalk detection models using satellite images suffer from low prediction accuracy since sidewalks occupy a very small portion of the visual information in satellite images compared to roads and buildings, are often obstructed by trees, bridges, and other urban structures, and their reduced area are particularly affected by sunlight condition~\cite{senlet2012segmentation, luo2019developing}.
A promising approach was using street-level images to tackle the obstruction issue; however, the resulting sidewalks were represented as polyline features, making them unsuitable for various analyses such as width measurement~\cite{ning2021sidewalk}.

To address these challenges, we introduce a method for detecting and morphological analysis of sidewalks from orthorectified aerial images. To overcome the high cost of pixel-wise annotation, we use publicly available planimetric data of New York City (NYC) (described in Section~\ref{sub:dataset}) to create accurate ground truth annotations for the obtained orthoimages. We train a multi-scale attention-based semantic segmentation model~\cite{tao2020hierarchical}  to detect roads, sidewalks, and buildings from aerial images. The specific architecture of the model enables detecting the sidewalks with very high precision, while the properties of the planimetric sidewalk data allow the model to make a correct prediction for various instances of the occluded sidewalks. The trained model is then employed to extract the sidewalk data from unlabeled images of Manhattan to be used for further analysis. 

The main contributions of this paper are: 
(1) A method that leverages a rich, publicly-available data set as the ground truth to train and test a recently developed semantic segmentation model capable of detecting sidewalks with high accuracy.
(2) Introduction of a robust approach to estimate meaningful sidewalk features (width, curvature, and angle) from the segmentation results.

\vspace{-0.3cm}
\section{Materials and method} \label{sec:method}
\vspace{-0.3cm}
 
To detect sidewalks from satellite images, we trained an attention-based semantic segmentation model using orthorectified images and annotation masks created from planimetric shapefiles of sidewalks, roads, and buildings.
The trained model is employed to extract the sidewalks from unlabeled images. Shape analysis algorithms are then applied to the extracted features to produce the indicators. To evaluate the performance of our approach, we also compute each indicator for the ground-truth annotation labels and validate our results against these extracted metrics. The proposed shape analysis method is summarized in Figure~\ref{fig:pipeline}.
We detail the main components of our approach next.

\vspace{-0.5cm}
\subsubsection{Data description.} \label{sub:dataset}

Raw satellite images have inherent distortions that cause feature displacement and scaling errors, resulting in inaccurate direct measurement of distance, angles, areas, and positions. Such distortions are corrected in the process of orthorectification to create accurately georeferenced images while preserving the distances between geographical features~\cite{tucker2004nasa}. 
Planimetric data are created from orthorectified aerial images due to their high accuracy in representing Earth's surface. Hence they are suitable choices for creating annotation masks. The majority of the NYC planimetric data is manually digitized~\cite{nycplanimetric}. To create the training set, we obtained 15,400 orthorectified tiles captured in 2018 from Manhattan and Brooklyn~\cite{nysortho}. For each tile, an annotation mask was created from planimetric sidewalks, roads, and buildings shapefiles clipped to the geographical extent of the tile. The image and annotation data were split into training (60\%), validation (20\%), and test sets (20\%). 

\begin{figure}[t!]
\centering
	\includegraphics[width=.75\linewidth]{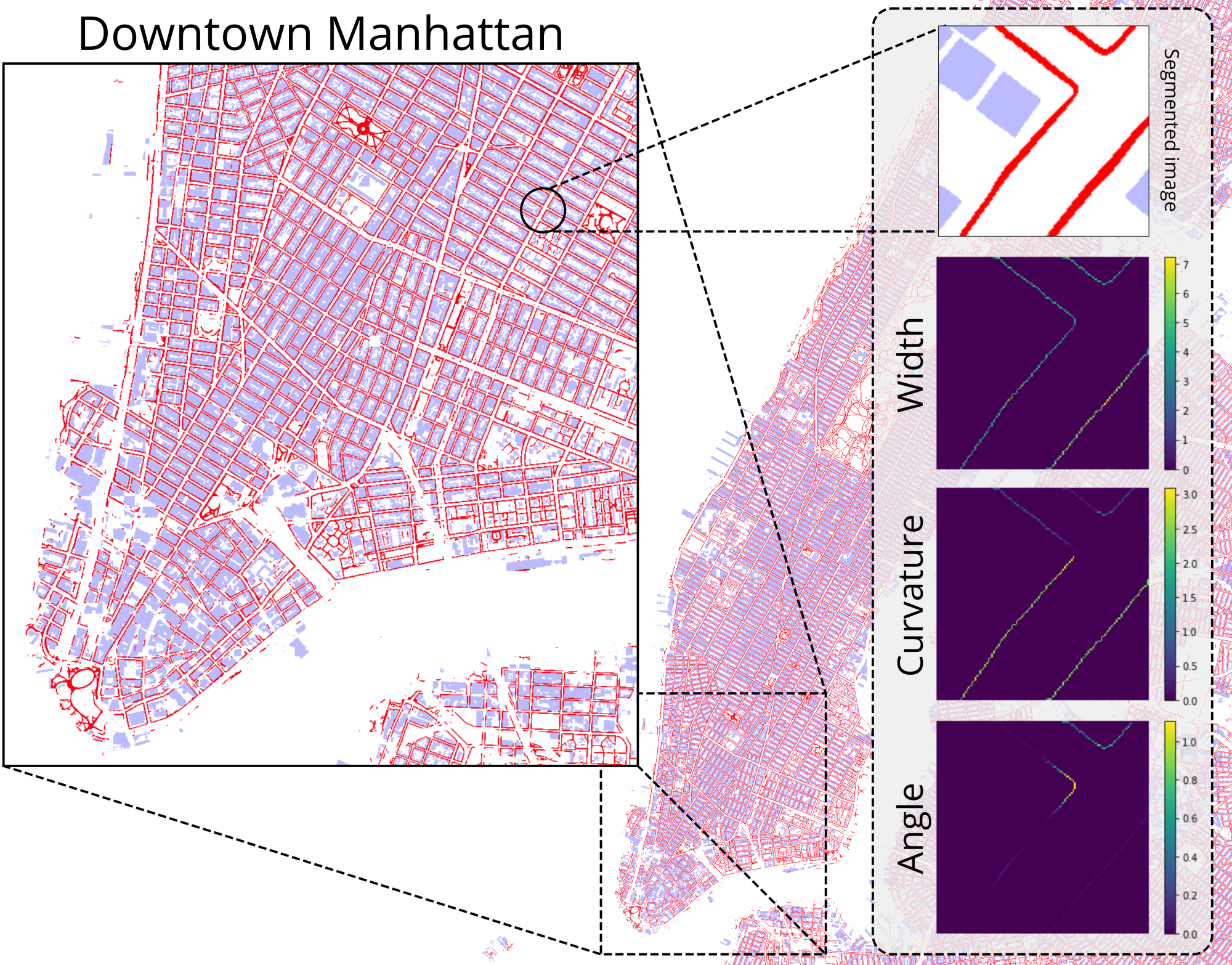}
	\caption{New York City segmented sidewalks (red) and buildings (blue). All other pixels are labeled white to highlight the sidewalks segmentation. The figure also highlights examples of the features extracted from a segmented tile. \label{fig:map}} 
	\vspace{-5mm}
\end{figure}
% \vspace{-0.75cm}

% \subsubsection{Method}
\vspace{-0.6cm}
\subsubsection{Semantic segmentation.} \label{sub:net_arch}

For the specific task at hand, we adopt the Hierarchical Multi-Scale Attention~\cite{tao2020hierarchical} with HRNet-OCR \cite{wang2020deep} backbone. HRNet connects high-to-low resolution convolutions via parallel and repeated multi-scale fusions to better preserve low-resolution representations alongside the high-resolution ones compared to previous works~\cite{yu2018deep} and has shown superior performance across segmentation benchmarks. The network is then trained for 200 epochs (batch size of 16); Rectified Adam as optimizer with polynomial learning rate policy. 

\vspace{-0.5cm}
\subsubsection{Sidewalk metrics.} \label{sub:metrics}
Using our trained model, we made predictions on the entire Manhattan's unlabeled data, which amounts to roughly 20000 tiles. The predictions are then used to estimate the sidewalks' width, angle, and curvature by employing geometry and image processing techniques~\cite{costabook09}. 
% In the adopted approach, we use a fundamental image processing step 
We use the \emph{skeletonization} technique to measure the attributes of interest. In colloquial terms, the skeleton of a binary shape is a thin line equidistant to the shape's borders. Here, the skeleton is obtained by a systematic sequence of morphological \emph{thinning} of the shape. For each point in the skeleton, we define the \emph{width} of the shape at a given point in the skeleton as twice the distance to the borders of the shape. The sidewalk \emph{angle} is estimated by its slope at a given point of the skeleton. A finite-difference approach is used to estimate this measurement % ~\cite{stoer2013introduction},
which relies on the determination of the orientation of the line connecting the query point to a neighboring one in the skeleton. The proposed approach is heavily dependent on the parameter $h$: the distance between the point of interest and its neighbor in the finite-difference operation, which defines the scale of the measurement. The sidewalk \emph{curvature} estimation considers the osculating circle passing through the query point and its two neighboring points in the skeleton~\cite{flynn1989reliable}. The curvature of a shape at the query point is defined as the inverse of the radius of the obtained circle.

\vspace{-0.3cm}
\section{Experimental results} \label{sec:experiments}
\vspace{-0.3cm}
\begin{wraptable}{r}{4.5cm}
\centering
\scriptsize
\vspace{-0.8cm}
\caption{Evaluation metrics.}
\vspace{-0.25cm}
\label{tab:netresults}
\begin{tabular}{lccc}
\toprule
Class & IoU & Precision & Recall \\ \midrule
Building & 0.83 & 0.92 & 0.90 \\
Road & 0.83 & 0.94 & 0.88 \\
Sidewalk & 0.79 & 0.91 & 0.86 \\
Background &0.86  & 0.92 & 0.93 \\ \midrule
\textbf{mIoU} & \multicolumn{3}{c}{0.83} \\ \bottomrule
\end{tabular}
\vspace{-0.75cm}
\end{wraptable}

Figure~\ref{fig:map} illustrates the segmentation and shape analysis results, and as can be seen, the model detected sidewalks and footpaths inside parks which are displayed in red. The model exhibits powerful detection capabilities in occluded areas since the annotation masks were based on the NYC planimetric data where sidewalks are mapped as continuous features even when occluded, given they are visible on both sides of the occluding element~\cite{nycplanimetric}. This property helped train the model to predict sidewalks correctly even when parts of them were occluded. 
 
\begin{wraptable}{l}{4.5cm}
\vspace{-0.75cm}
    \centering
    \scriptsize
    \caption{Measurements.}
    \vspace{-0.25cm}
    \label{tab:diffmeas}
    \begin{tabular}{@{}lcccrr@{}}
        \toprule
        Feature & \multicolumn{2}{c}{Bin} & N & RMSE \\
        \midrule
        \multirow{2}{*}{Width}& 0&7 & 1291 & 3.12 \\
        \multirow{2}{*}{(pixels)}& 7&14 & 11045 & 1.85 \\
        & 14&{\footnotesize $+\infty$} & 529 & 6.07 \\
        \midrule
        \multirow{2}{*}{Angle}& 0&45 & 539 & 0.84 \\
        \multirow{2}{*}{(degrees)}& 45&90 & 6599 & 0.21 \\
        & 90&135 & 4167 & 0.34 \\
        & 135&180 &  1560 & 0.66 \\
        \midrule
        \multirow{2}{*}{Curvature}&  0.0&0.1 & 4069 & 0.09 \\
        \multirow{2}{*}{(pixels)} & 0.1&0.2 & 2926 & 0.10 \\
         & 0.2&0.3 & 2688 & 0.10 \\
         & 0.3&{\footnotesize $+\infty$} & 3182 & 0.21 \\
        \bottomrule
    \end{tabular}
\vspace{-0.5cm}
\end{wraptable}

Table~\ref{tab:netresults} shows the performance metrics of our semantic segmentation model on the held-out test set.  Average and class-wise Intersection over Union (IoU) and precision and recall are presented. Sidewalks were detected with 79\% IoU, and the average IoU (mIoU) across all four classes is 83\%. Good results were attained in the identification not only of sidewalks but of all considered classes.

To further assess the sidewalks' condition, we measure three different average metrics: width, angle, and curvature as described in Section~\ref{sec:method}. The average of each measurement is calculated across all the values of a given image and is used as a proxy for the overall characterization of the sidewalks visible in each image. 
The average width, angle, and curvature are computed independently for the ground truth and prediction masks. The values are partitioned into bins, and the root mean squared error is calculated for each measurement's bin (Table~\ref{tab:diffmeas}). The RMSE values vary significantly across the bins: wider sidewalks, sidewalks with angles between 0-45$^\circ$, and the highly curved ones exhibit higher errors than the rest. 

\begin{wraptable}{r}{5.5cm}
\vspace{-0.8cm}
    \centering
    \scriptsize
    \caption{Aggregation by LU.}
    \vspace{-0.25cm}
    \label{tab:landuse}
    \begin{tabular}{@{}lcccc@{}}
        \toprule
        Land use (LU) & \# Img. & Width & Angle & Curv. \\
        \midrule
        % - & 39 & 5.595 & 1.307 & 0.254 \\
        Residential & 4530 & 8.91 & 85.07 & 0.175 \\
        Commercial & 1806 & 9.25 & 88.45 & 0.198 \\
        Industrial & 75 & 7.83 & 83.9 & 0.156 \\
        % Industrial-Transportation & 10 & 8.878 & 1.589 & 0.136 \\
        % Mixed & 5054 & 9.208 & 1.578 & 0.200 \\
        Public facilities & 934 & 8.54 & 91.9 & 0.20 \\
        % Parking & 96 & 8.310 & 1.485 & 0.194 \\
        Parks & 1904 & 8.56 & 99.2 & 0.27 \\
        % Parks/openspace-Commercial & 111 & 9.441 & 1.403 & 0.208 \\
        % Parks / mixed & 531 & 9.430 & 1.768 & 0.215 \\
        % Transportation & 488 & 6.331 & 1.319 & 0.142 \\
        % Vacant & 156 & 7.972 & 1.327 & 0.161 \\
        \bottomrule
    \end{tabular}
\vspace{-0.75cm}
\end{wraptable}
We also analyzed the relationship between the sidewalk attributes and land use at the tile level, calculated by doing a spatial join between the MapPLUTO data~\cite{pluto} and the tile extents. Table~\ref{tab:landuse} shows the three sidewalk measurements aggregated by land use. Commercial areas have the highest width, in line with the NYC design guides. As expected, parks exhibit the highest angle and curvature due to various bending pathways. Residential areas have the second lowest curvature suitable for the navigation of people with different mobility levels.

\vspace{-0.4cm}
\section{Conclusion} \label{sec:conclusions}
\vspace{-0.35cm}

We present a method for the semantic segmentation of sidewalks from orthorectified images of NYC with highly dense urban areas and prevalent cases of sidewalk occlusion. The method shows promising detection power, achieving 83\% mIoU over four classes with sidewalks having 79\% IoU and 91\% precision. Since the ground truth mapped sidewalks regardless of the shadow and occlusions, the evaluation metrics on the held-out test sets of 3000 images indicate that the model performed well in predicting sidewalks under different conditions, including occlusion cases. The trained model was used to extract the sidewalks, roads, and buildings for the whole of Manhattan. Morphological and discrete geometry operations were then applied to calculate the sidewalks' width, angle, and curvature for the ground-truth and model's predictions. The error rate for each measurement was evaluated per range of values. Finally, the sidewalk metrics were aggregated per land use. 

The results show promising potentials of the method, offering several exciting opportunities for future work. We plan to aggregate the measurements on the sidewalk segment level to provide a more informative overview of the sidewalks' condition. We also plan to create a more generalizable model by expanding the training data to include cities with varying topological characteristics and shadow patterns~\cite{8283638}.

% \newpage
\vspace{-0.3cm}
\subsubsection{Acknowledgments.}
\vspace{-0.3cm}

This work has been partially funded by C2SMART, São Paulo Research Foundation (FAPESP) grants \#2015/22308-2 and \#2019/01077-3,  NSF awards CNS-1229185, CCF-1533564, CNS-1544753, CNS-1730396, and CNS-1828576. Silva is partially funded by DARPA. Any opinions, findings, and conclusions or recommendations expressed in this material are those of the authors and do not necessarily reflect the views of DARPA.

% \vspace{-1cm}
\bibliographystyle{splncs04}
\bibliography{references}

\begin{thebibliography}{10}
\providecommand{\url}[1]{\texttt{#1}}
\providecommand{\urlprefix}{URL }
\providecommand{\doi}[1]{https://doi.org/#1}

\bibitem{beale2006mapping}
Beale, L., Field, K., Briggs, D., Picton, P., Matthews, H.: Mapping for
  wheelchair users: Route navigation in urban spaces. The Cartographic Journal
  \textbf{43}(1),  68--81 (2006)

\bibitem{costabook09}
Costa, L.F., Cesar-Jr., R.: Shape Classification and Analysis: Theory and
  Practice. Taylor \& Francis (2009)

\bibitem{deitz2021squeaky}
Deitz, S., Lobben, A., Alferez, A.: Squeaky wheels: Missing data, disability,
  and power in the smart city. Big Data \& Society  \textbf{8}(2) (2021)

\bibitem{flynn1989reliable}
Flynn, P.J., Jain, A.K.: On reliable curvature estimation. In: Proc. of the
  IEEE Conf. on Comp. Vision and Pattern Recognition (1989)

\bibitem{luo2019developing}
Luo, J., Wu, G., Wei, Z., Boriboonsomsin, K., Barth, M.: Developing an
  aerial-image-based approach for creating digital sidewalk inventories.
  Transportation research record  \textbf{2673}(8),  499--507 (2019)

\bibitem{8283638}
Miranda, F., Doraiswamy, H., Lage, M., Wilson, L., Hsieh, M., Silva, C.T.:
  Shadow accrual maps: Efficient accumulation of city-scale shadows over time.
  IEEE Transactions on Visualization and Computer Graphics  \textbf{25}(3),
  1559--1574 (2019)

\bibitem{mosaic}
Miranda, F., Hosseini, M., Lage, M., Doraiswamy, H., Dove, G., Silva, C.T.:
  {Urban Mosaic}: Visual exploration of streetscapes using large-scale image
  data. In: Proc. of the 2020 CHI Conference on Human Factors in Computing
  Systems (2020)

\bibitem{ning2021sidewalk}
Ning, H., Ye, X., Chen, Z., Liu, T., Cao, T.: Sidewalk extraction using aerial
  and street view images. Environment and Planning B: Urban Analytics and City
  Science  (2021)

\bibitem{pluto}
{NYC DCP}: {PLUTO and MapPLUTO}. Available:
  \url{https://www1.nyc.gov/site/planning/data-maps/open-data/dwn-pluto-mappluto.page}

\bibitem{nycplanimetric}
{NYC DoITT}: {New York City Planimetrics Data}. Available:
  \url{https://github.com/CityOfNewYork/nyc-planimetrics}

\bibitem{nysortho}
{NYC GIS}: {NYS Statewide Digital Orthoimagery Program}. Available:
  \url{https://gis.ny.gov/gateway/orthoprogram/index.cfm}

\bibitem{osama2017evaluating}
Osama, A., Sayed, T.: Evaluating the impact of connectivity, continuity, and
  topography of sidewalk network on pedestrian safety. Accident Analysis \&
  Prevention  \textbf{107},  117--125 (2017)

\bibitem{rundle2011using}
Rundle, A.G., Bader, M.D., Richards, C.A., Neckerman, K.M., Teitler, J.O.:
  Using {Google Street View} to audit neighborhood environments. American
  Journal of Preventive Medicine  \textbf{40}(1),  94--100 (2011)

\bibitem{saha2019project}
Saha, M., Saugstad, M., Maddali, H.T., Zeng, A., Holland, R., Bower, S., Dash,
  A., Chen, S., Li, A., Hara, K., Froehlich, J.: Project sidewalk: A web-based
  crowdsourcing tool for collecting sidewalk accessibility data at scale. In:
  Proc. of the 2019 CHI Conference on Human Factors in Computing Systems (2019)

\bibitem{senlet2012segmentation}
Senlet, T., Elgammal, A.: Segmentation of occluded sidewalks in satellite
  images. In: Proc. of the 21st Int. Conf. on Pattern Recognition. IEEE (2012)

\bibitem{tao2020hierarchical}
Tao, A., Sapra, K., Catanzaro, B.: Hierarchical multi-scale attention for
  semantic segmentation. arXiv preprint arXiv:2005.10821  (2020)

\bibitem{tucker2004nasa}
Tucker, C.J., Grant, D.M., Dykstra, J.D.: Nasa’s global orthorectified
  landsat data set. Photogrammetric Engineering \& Remote Sensing
  \textbf{70}(3),  313--322 (2004)

\bibitem{wang2020deep}
Wang, J., Sun, K., Cheng, T., Jiang, B., Deng, C., Zhao, Y., Liu, D., Mu, Y.,
  Tan, M., Wang, X., Liu, W., Xiao, B.: Deep high-resolution representation
  learning for visual recognition. IEEE Trans. on Pattern Analysis \& Machine
  Intelligence  (2020)

\bibitem{yu2018deep}
Yu, F., Wang, D., Shelhamer, E., Darrell, T.: Deep layer aggregation. In: Proc.
  of the IEEE Conf. on Comp. Vision and Pattern Recognition (2018)

\end{thebibliography}
\end{document}